\documentclass[10pt,twocolumn,letterpaper]{article}

\usepackage{cvpr}
\usepackage{times}
\usepackage{epsfig}
\usepackage{graphicx}
\usepackage{amsmath}
\usepackage{amssymb}
\usepackage{multirow}
\usepackage{subcaption}

\usepackage[pagebackref=true,breaklinks=true,colorlinks,bookmarks=false]{hyperref}

\cvprfinalcopy 

\def\cvprPaperID{****} 

\begin{document}

\title{Bias in Multimodal AI: Testbed for Fair Automatic Recruitment}

\author{Alejandro Pe\~na, Ignacio Serna, Aythami Morales, Julian Fierrez\\
School of Engineering, Universidad Autonoma de Madrid, Spain\\
{\tt\small \{alejandro.penna, ignacio.serna.es, aythami.morales,  julian.fierrez\}@uam.es}
}

\maketitle

\begin{abstract}
The presence of decision-making algorithms in society is rapidly increasing nowadays, while concerns about their transparency and the possibility of these algorithms becoming new sources of discrimination are arising. In fact, many relevant automated systems have been shown to make decisions based on sensitive information or discriminate certain social groups (e.g. certain biometric systems for person recognition). With the aim of studying how current multimodal algorithms based on heterogeneous sources of information are affected by sensitive elements and inner biases in the data, we propose a fictitious automated recruitment testbed: FairCVtest. We train automatic recruitment algorithms using a set of multimodal synthetic profiles consciously scored with gender and racial biases. FairCVtest shows the capacity of the Artificial Intelligence (AI) behind such recruitment tool to extract sensitive information from unstructured data, and exploit it in combination to data biases in undesirable (unfair) ways. Finally, we present a list of recent works developing techniques capable of removing sensitive information from the decision-making process of deep learning architectures. We have used one of these algorithms (SensitiveNets) to experiment discrimination-aware learning for the elimination of sensitive information in our multimodal AI framework. Our methodology and results show how to generate fairer AI-based tools in general, and in particular fairer automated recruitment systems.
\end{abstract}

\section{Introduction}\label{Introduction}
Over the last decades we have witnessed great advances in fields such as data mining, Internet of Things, or Artificial Intelligence, among others, with data taking on special relevance. Paying particular attention to the field of machine learning, the large amounts of data currently available have led to a paradigm shift, with handcrafted algorithms being replaced in recent years by deep learning technologies. 




Machine learning algorithms rely on data collected from society, and therefore may reflect current and historical biases \cite{BigDataImpact} if appropriate measures are not taken. In this scenario, machine learning models have the capacity to replicate, or even amplify human biases present in the data \cite{acien2018bias,demographic_bias_biometric,DL_prejudiced,MenAlsoLike}. There are relevant models based on machine learning that have been shown to make decisions largely influenced by gender or ethnicity. Google's \cite{Disc_Google} or Facebook's \cite{Disc_facebook} ad delivery systems generated undesirable discrimination with disparate performance across population groups \cite{Gender_shades}. New York’s insurance regulator probed UnitedHealth Group over its use of an algorithm that researchers found to be racially biased, the algorithm prioritized healthier white patients over sicker black ones \cite{wallstreet2019bias}. More recently,  Apple Credit service granted higher credit limits to men than women\footnote{\url{https://edition.cnn.com/2019/11/10/business/goldman-sachs-apple-card-discrimination/}} even though it was programmed to be blind to that variable (the biased results in this case were originated from other variables \cite{apple2019bias}).

The usage of AI is also growing in human resources departments, with video- and text-based screening software becoming increasingly common in the hiring pipeline \cite{video_interview}. But automatic tools in this area have exhibited worrying biased behaviors in the past. For example, Amazon's recruiting tool was preferring male candidates over female candidates \cite{amazon2018bias}. The access to better job opportunities is crucial to overcome differences of minority groups. However, in cases such as automatic recruitment, both the models and their training data are usually private for corporate or legal reasons. This lack of transparency, along with the long history of bias in the hiring domain, hinder the technical evaluation of these systems in search of possible biases targeting protected groups \cite{Bias_in_Hiring}.

\begin{figure*}[t!]
\centering
\includegraphics[width = 0.8\textwidth]{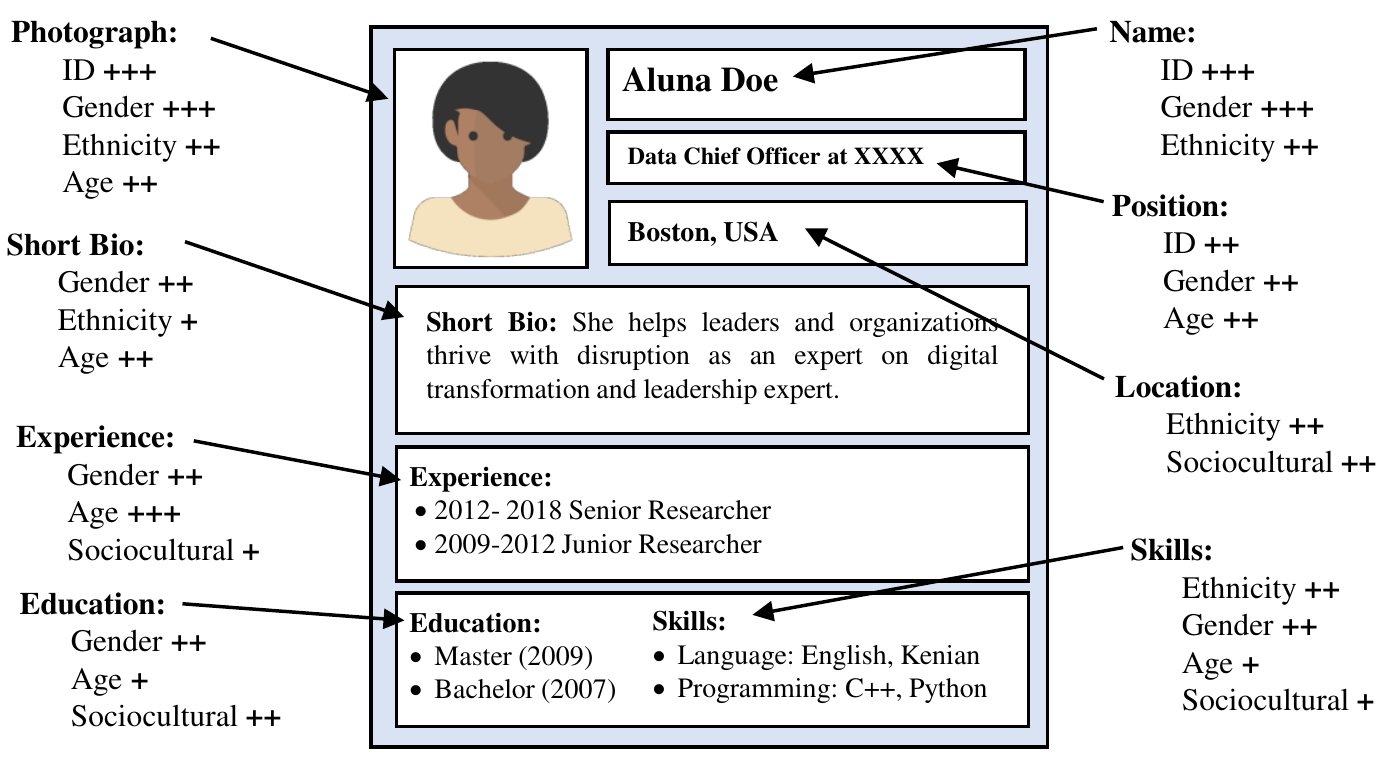}
\caption{Information blocks in a resume and personal attributes that can be derived from each one. The number of crosses represent the level of sensitive information (+++ =  high, ++ = medium, + = low).}
\label{cv}
\end{figure*}

This deployment of automatic systems has led governments to adopt regulations in this matter, placing special emphasis on personal data processing and preventing algorithmic discrimination. Among these regulations, the new European Union's General Data Protection Regulation (GDPR)\footnote{\url{https://gdpr.eu/}}, adopted in May $2018$, is specially relevant for its impact on the use of machine learning algorithms \cite{Right_explanation}. The GDPR aims to protect EU citizens' rights concerning data protection and privacy by regulating how to collect, store, and process personal data (e.g. Articles $17$ and $44$). This normative also regulates the ``right to explanation" (e.g. Articles $13$-$15$), by which citizens can ask for explanations about algorithmic decisions made about them, and requires measures to prevent discriminatory effects while processing sensitive data (according to Article $9$, sensitive data includes ``personal data revealing racial or ethnic origin, political opinions, religious or philosophical beliefs").


On the other hand, one of the most active areas in machine learning is around the development of new multimodal models capable of understanding and processing information from multiple heterogeneous sources of information \cite{Baltruaitis2017MultimodalML}. Among such sources of information we can include structured data (e.g. in tables), and unstructured data from images, audio, and text. The implementation of these models in society must be accompanied by effective measures to prevent algorithms from becoming a source of discrimination. In this scenario, where multiple sources of both structured and unstructured data play a key role in algorithms' decisions, the task of detecting and preventing biases becomes even more relevant and difficult.

In this environment of desirable fair and trustworthy AI, the main contributions of this work are:
\begin{itemize}
\setlength\itemsep{-0.5em}
    \item We present a new public experimental framework around automated recruitment aimed to study how multimodal machine learning is influenced by biases present in the training datasets: FairCVtest\footnote{\url{https://github.com/BiDAlab/FairCVtest}}.
    \item We have evaluated the capacity of popular neural network to learn biased target functions from multimodal sources of information including images and structured data from resumes. 
    \item We develop a discrimination-aware learning method based on the elimination of sensitive information such as gender or ethnicity from the learning process of multimodal approaches, and apply it to our automatic recruitment testbed for improving fairness.
\end{itemize}

Our results demonstrate the high capacity of commonly used learning methods to expose sensitive information (e.g. gender and ethnicity) and the necessity to implement appropriate techniques to guarantee discrimination-free decision-making processes. 

The rest of the paper is structured as follows: Section~\ref{Framework} analyzes the information available in a typical resume and the sensitive data associated to it. Section~\ref{Problem_formulation} presents the general framework for our work including problem formulation and the dataset created in this work: FairCVdb. Section~\ref{experiments} reports the experiments in our testbed FairCVtest after describing the experimental methodology and the different scenarios evaluated. Finally, Section~\ref{conclusions} summarizes the main conclusions.




\begin{figure*}[t]
\centering
\includegraphics[width=0.75\textwidth]{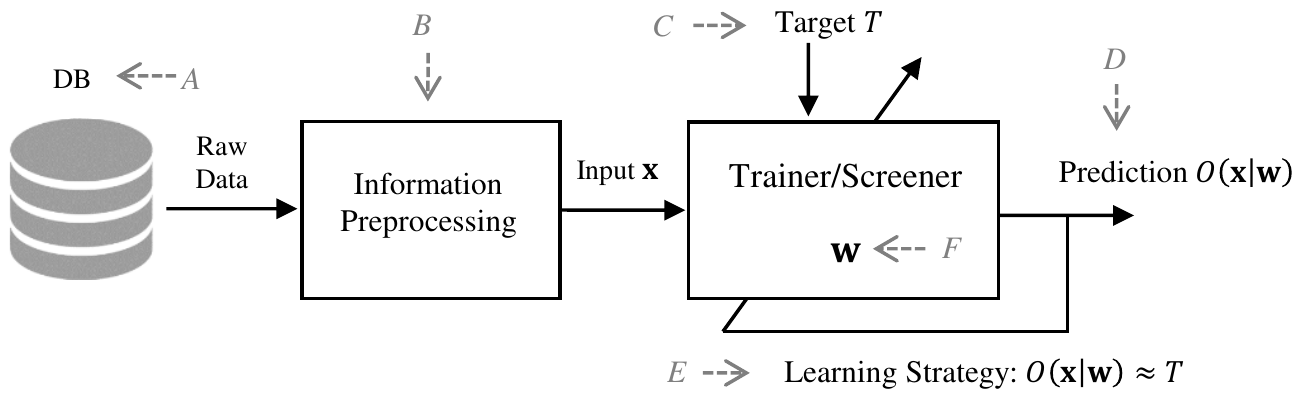} 
\caption{Block diagram of the automatic learning process and $6$ (\emph{A} to \emph{E}) stages where bias can appear.}
\label{Block_diagram}
\end{figure*}

\section{What else does your resume data reveal? Studying multimodal biases in AI }\label{Framework}

For the purpose of studying discrimination in Artificial Intelligence at large, in this work we propose a new experimental framework inspired in a fictitious automated recruiting system: FairCVtest.

There are many companies that have adopted predictive tools in their recruitment processes to help hiring managers find successful employees. Employers often adopt these tools in an attempt to reduce the time and cost of hiring, or to maximize the quality of the hiring process, among other reasons \cite{Hiring_Algorithms_Report}. We chose this application because it comprises personal information from different nature \cite{2018_INFFUS_MCSreview1_Fierrez}. 

The resume is traditionally composed by structured data including name, position, age, gender, experience, or education, among others (see Figure \ref{cv}), and also includes unstructured data such as a face photo or a short biography. A face image is rich in unstructured information such as identity, gender, ethnicity, or age \cite{2018_TIFS_SoftWildAnno_Sosa,Understanding_faces}. That information can be recognized in the image, but it requires a cognitive or automatic process trained previously for that task. The text is also rich in unstructured information. The language and the way we use that language, determine attributes related to your nationality, age, or gender. Both, image and text, represent two of the domains that have attracted major interest from the AI research community during last years. The Computer Vision and the Natural Language Processing communities have boosted the algorithmic capabilities in image and text analysis through the usage of massive amounts of data, large computational capabilities (GPUs), and deep learning techniques. 

The resumes used in the proposed FairCVtest framework include merits of the candidate (e.g. experience, education level, languages,  etc...), two demographic attributes (gender and ethnicity), and a face photograph (see Section \ref{Dataset} for all the details). 

\section{Problem formulation and dataset}
\label{Problem_formulation}

The model represented by its parameters vector $\textbf{w}$ is trained according to multimodal input data defined by $n$ features $\textbf{x} = [x_1,...,x_n] \in \mathbb{R}^n$, a Target function $T$, and a learning strategy that minimizes the error between the output $O$ and the Target function $T$. In our framework where $\textbf{x}$ is data obtained from the resume, $T$ is a score within the interval [$0$, $1$] ranking the candidates according to their merits. A score close to $0$ corresponds to the worst candidate, while the best candidate would get $1$. Biases can be introduced in different stages of the learning process (see Figure \ref{Block_diagram}): in the Data used to train the models (\textit{A}), the Preprocessing or feature selection (\textit{B}), the Target function (\textit{C}), and the Learning strategy (\textit{E}). As a result, a biased Model (\textit{F}) will produce biased Results (\textit{D}). In this work we focus on the Target function (\textit{C}) and the Learning strategy (\textit{E}). The Target function is critical as it could introduce cognitive biases from biased processes. The Learning strategy is traditionally based on the minimization of a loss function defined to obtain the best performance. The most popular approach for supervised learning is to train the model $\textbf{w}$ by minimizing a loss function $\mathcal{L}$ over a set of training samples $\mathcal{S}$:

\begin{equation}
\label{eqn:learning_strategy}
    \min_{\textbf{w}}{\sum_{\textbf{x}^{j} \in \mathcal{S}}\mathcal{L}(O(\textbf{x}^j|\textbf{w}),T^j)} 
\end{equation}


\subsection{FairCVdb: research dataset for multimodal AI}
\label{Dataset}
We have generated $24$,$000$ synthetic resume profiles including $12$ features obtained from $5$ information blocks, $2$ demographic attributes (gender and ethnicity), and a face photograph. The $5$ blocks are: $1$) education attainment (generated from US Census Bureau $2018$ Education Attainment data\footnote{\url{https://www.census.gov/data/tables/2018/demo/education-attainment/cps-detailed-tables.html}}, without gender or ethnicity distinction), $2$) availability, $3$) previous experience, $4$) the existence of a recommendation letter, and $5$) language proficiency in a set of $8$ different and common languages (chosen from US Census Bureau Language Spoken at Home data\footnote{\url{https://www.census.gov/data/tables/2013/demo/2009-2013-lang-tables.html}}). Each language is encoded with an individual feature ($8$ features in total) that represents the level of knowledge in that language.

\begin{figure}[t!]
\centering
\includegraphics[width=1\columnwidth]{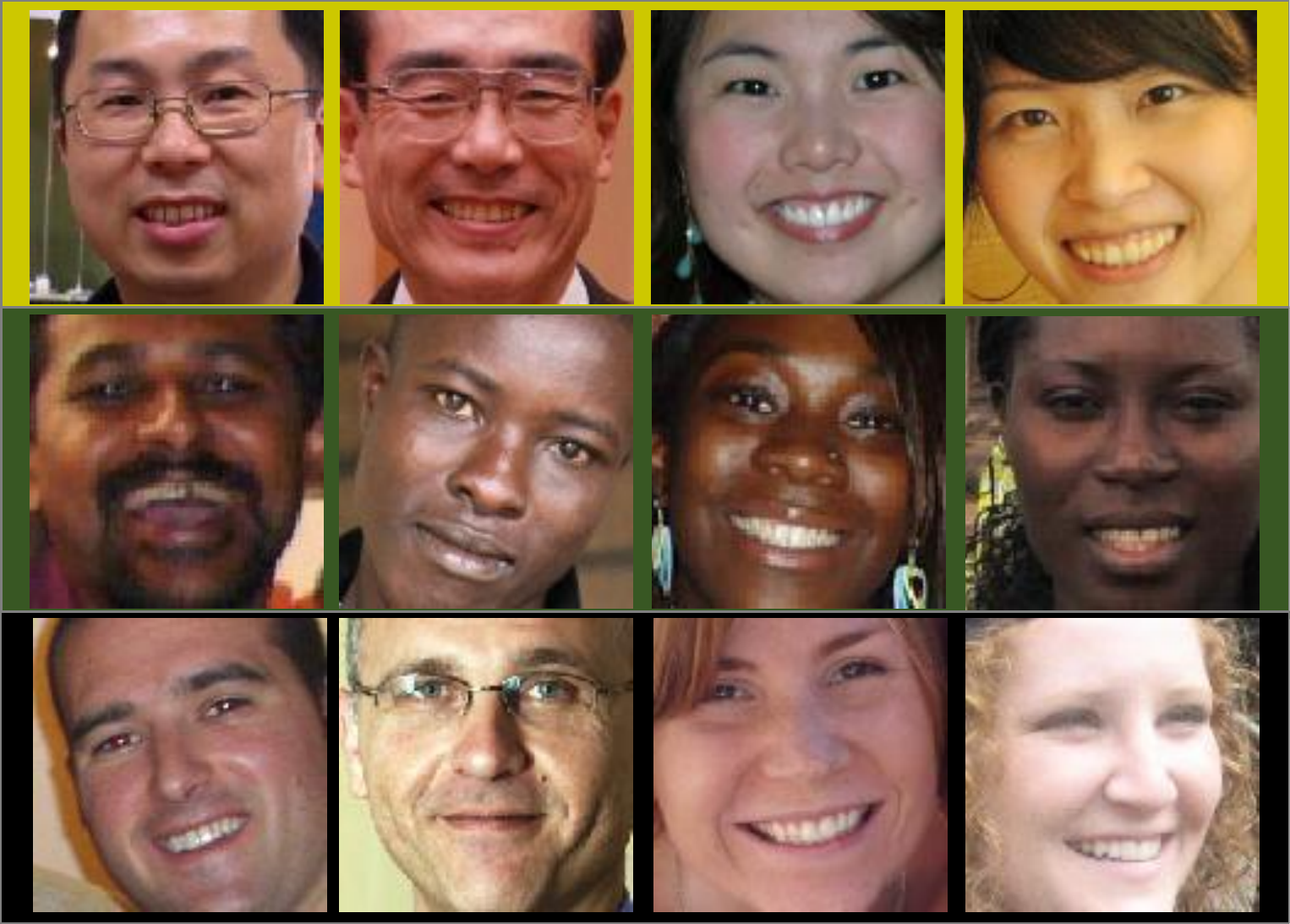} 
\caption{Examples of the six demographic groups included in DiveFace: male/female for $3$ ethnic groups.}
\label{diveface}
\end{figure}

Each profile has been associated according to the gender and ethnicity attributes with an identity of the DiveFace database \cite{SensitiveNets}. DiveFace contains face images ($120 \times 120$ pixels) and annotations equitably distributed among $6$ demographic classes related to gender and $3$ ethnic groups (Black, Asian, and Caucasian), including $24$K different identities (see Figure \ref{diveface}). 

Therefore, each profile in FairCVdb includes information on gender and ethnicity, a face image (correlated with the gender and ethnicity attributes), and the $12$ resume features described above, to which we will refer to candidate competencies $x_i$. 

\begin{figure*}[t]
\centering
\includegraphics[width=1\textwidth]{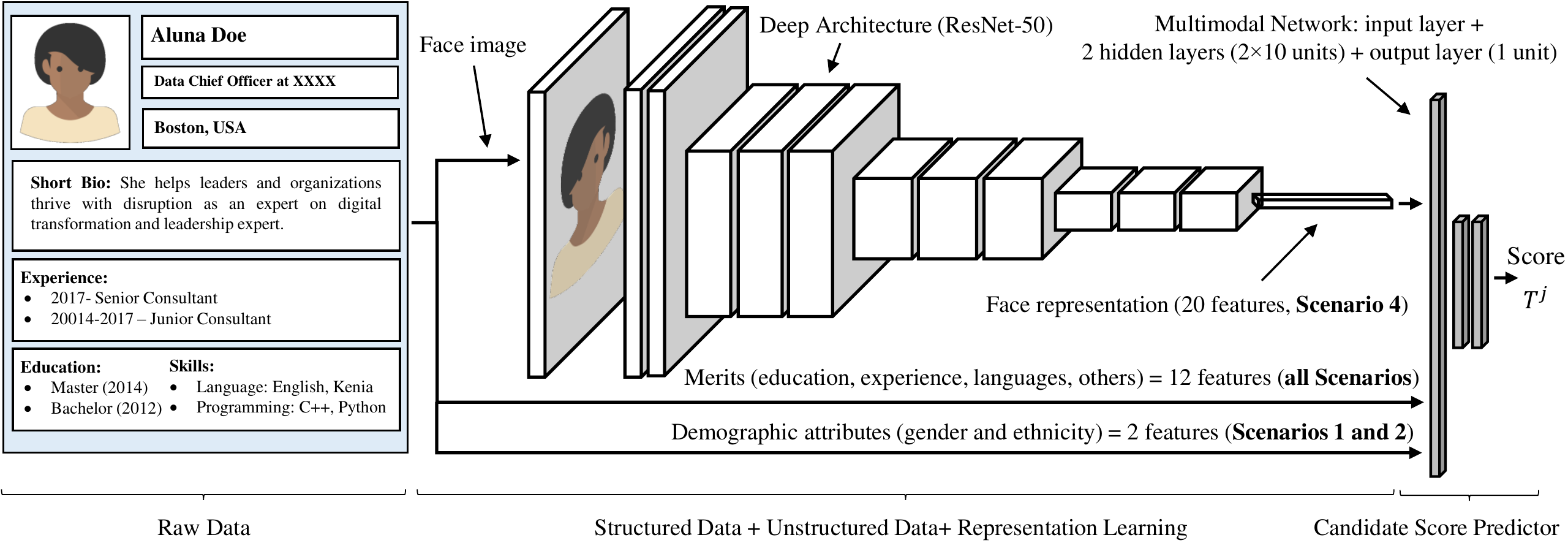} 
\caption{Multimodal learning architecture composed by a Convolutional Neural Network (ResNet-$50$) and a fully connected network used to fuse the features from different domains (image and structured data). Note that some features are included or removed from the learning architecture depending of the scenario under evaluation.}
\label{network}
\end{figure*}

The score $T^j$ for a profile $j$ is generated by linear combination of the candidate competencies $\textbf{x}^j = [x^j_1, ..., x^j_n]$ as:
\begin{equation}
\label{eqn:Score_gen}
    T^j = \beta^j + \sum_{i = 1}^{n} \alpha_{i} x^j_i 
\end{equation}
where $n = 12$ is the number of features (competencies), $\alpha_i$ are the weighting factors for each competency $x_i^j$ (fixed manually based on consultation with a human recruitment expert), and $\beta^j$ is a small Gaussian noise to introduce a small degree of variability (i.e. two profiles with the same competencies do not necessarily have to obtain the same result in all cases). Those scores $T^j$ will serve as groundtruth in our experiments.

Note that, by not taking into account gender or ethnicity information during the score generation in Equation~(\ref{eqn:Score_gen}), these scores become agnostic to this information, and should be equally distributed among different demographic groups. Thus, we will refer to this target function as Unbiased scores $T^U$, from which we define two target functions that include two types of bias: Gender bias $T^G$ and Ethnicity bias $T^E$. Biased scores are generated by applying a penalty factor $T_\delta$ to certain individuals belonging to a particular demographic group. This leads to a set of scores where, with the same competencies, certain groups have lower scores than others, simulating the case where the process is influenced by certain cognitive biases introduced by humans, protocols, or automatic systems.

\section{FairCVtest: Description and experiments}\label{experiments}

\subsection{FairCVtest: Scenarios and protocols}\label{protocol}
In order to evaluate how and to what extent an algorithm is influenced by biases that are present in the FairCVdb target function, we use the FairCVdb dataset previously introduced in Section \ref{Problem_formulation} to train various recruitment systems under different scenarios. The proposed FairCVtest testbed consist of FairCVdb, the trained recruitment systems, and the related experimental protocols.

First, we present $4$ different versions of the recruitment tool, with slight differences in the input data and target function aimed at studying different scenarios concerning gender bias. After that, we will show how those scenarios can be easily extrapolated to ethnicity bias. 

The $4$ Scenarios included in FairCVtest were all trained using the competencies presented on Section \ref{Problem_formulation}, with the following particular configurations:

\begin{itemize}
\setlength\itemsep{-0.5em}
    \item \textit{Scenario 1:} Training with Unbiased scores $T^U$, and the gender attribute as additional input.
    \item \textit{Scenario 2:} Training with Gender-biased scores
    $T^G$, and the gender attribute as additional input.
    \item \textit{Scenario 3:} Training with Gender-biased scores $T^G$, but the gender attribute wasn't given as input.
    \item \textit{Scenario 4:} Training with Gender-biased scores
    $T^G$, and a feature embedding from the face photograph as additional input.
\end{itemize}

In all $4$ cases, we designed the candidate score predictor as a feedforward neural network with two hidden layers, both of them composed by $10$ neurons with ReLU activation, and only one neuron with sigmoid activation in the output layer, treating this task as a regression problem.

In Scenario $4$, where the system takes also as input an embedding from the applicant's face image, we use the pretrained model ResNet-$50$ \cite{he2015deep} as feature extractor to obtain these embeddings. ResNet-$50$ is a popular Convolutional Neural Network, originally proposed to perform face and image recognition, composed with $50$ layers including residual or ``shortcuts" connections to improve accuracy as the net depth increases. ResNet-$50$'s last convolutional layer outputs embeddings with $2048$ features, and we added a fully connected layer to perform a bottleneck that compresses these embeddings to just $20$ features (maintaining competitive face recognition performances). Note that this face model was trained exclusively for the task of face recognition. Gender and ethnicity information were not intentionally employed during the training process. Of course, this information is part of the face attributes.

Figure \ref{network} summarizes the general learning architecture of FairCVtest. The experiments performed in next section will try to evaluate the capacity of the recruitment AI to detect protected attributes (e.g. gender, ethnicity) without being explicitly trained for this task.

\subsection{FairCVtest: Predicting the candidate score}\label{bias}

\begin{figure}[t]
\centering
\includegraphics[width=\columnwidth]{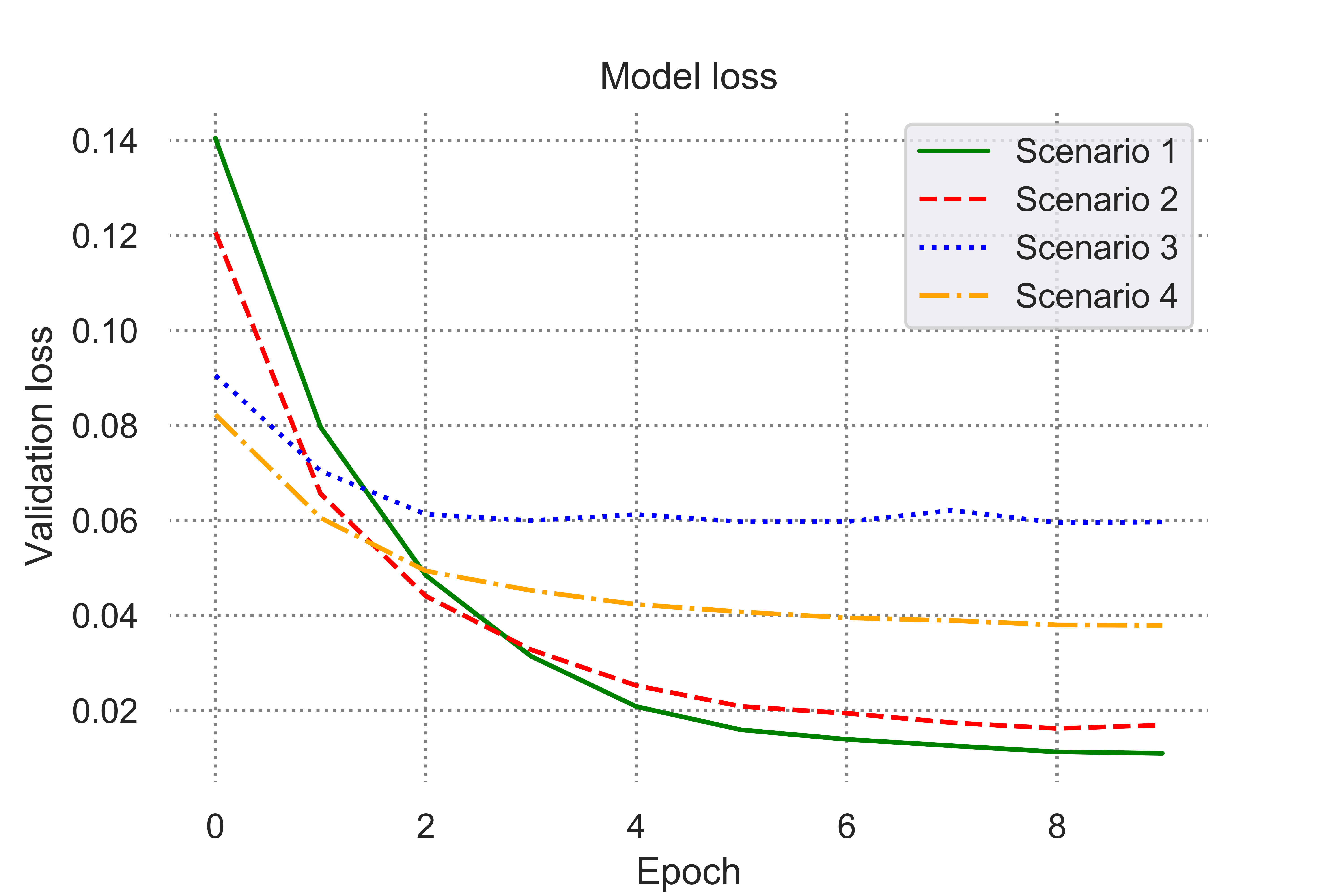} 
\caption{Validation loss during the training process obtained for the different scenarios.}
\label{val_loss}
\end{figure}

\begin{figure*}[t]
\centering
\includegraphics[width=0.77\textwidth]{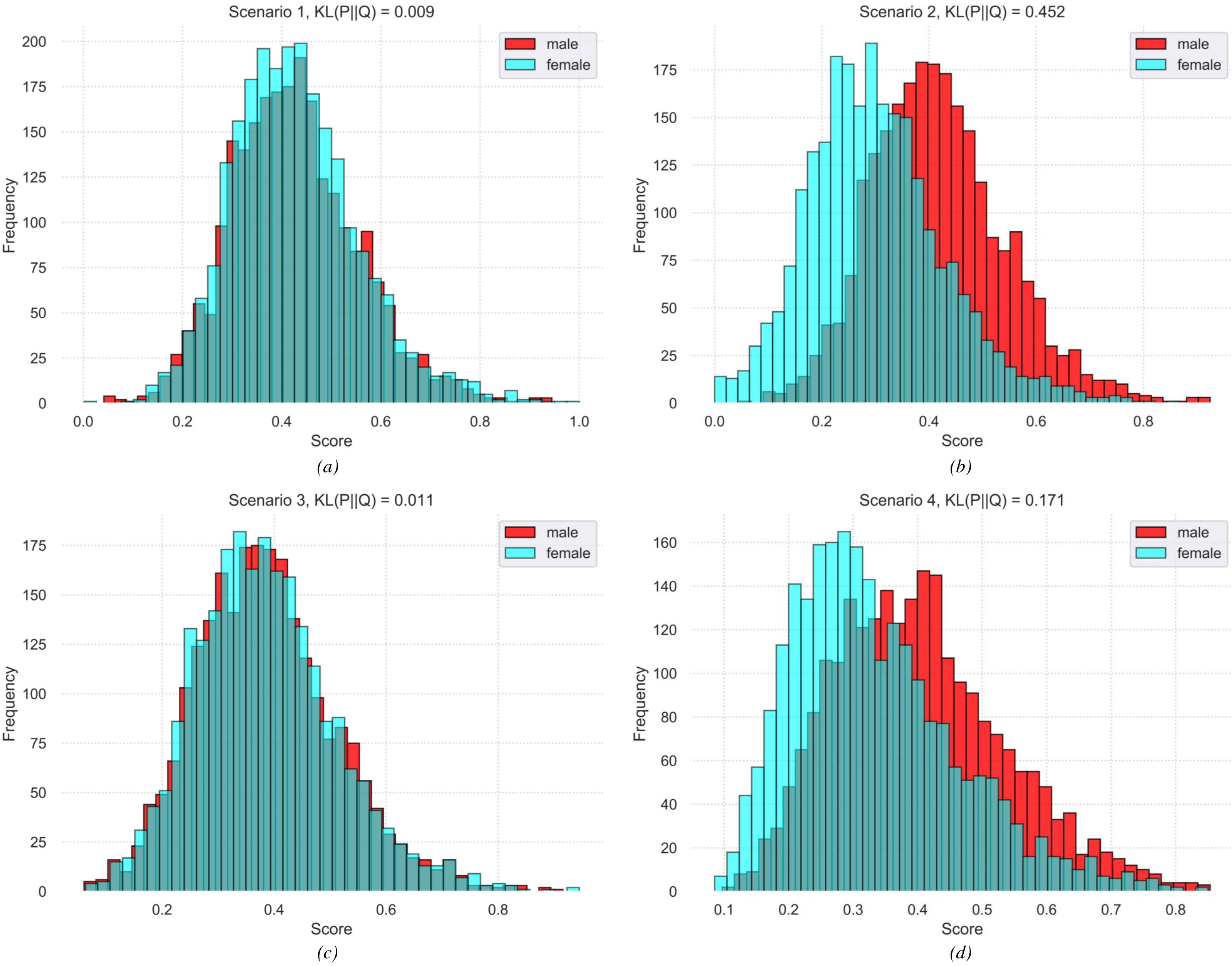}
\caption{Hiring score distributions by gender for each Scenario. The results show how multimodal learning is capable to reproduce the biases present in the training data even if the gender attribute is not explicitly available.}
\label{gender_distribution}
\end{figure*}

The recruitment tool was trained with the $80$\% of the synthetic profiles ($19$,$200$ CVs) described in Section \ref{Dataset}, and retaining $20$\% as validation set ($4$,$800$ CVs), each set equally distributed among gender and ethnicity, using Adam optimizer, $10$ epochs, batch size of $128$, and mean absolute error as loss metric. 

In Figure \ref{val_loss} we can observe the validation loss during the training process for each Scenario (see Section~\ref{protocol}), which gives us an idea about the performance of each network in the main task (i.e. scoring applicants' resumes). In the first two scenarios the network is able to model the target function more precisely, because in both cases it has all the features that influenced in the score generation. Note that, by adding a small Gaussian noise to include some degree of variability, see Equation~(\ref{eqn:Score_gen}), this loss will never converge to $0$. Scenario $3$ shows the worst performance, what makes sense since there's no correlation between the bias in the scores and the inputs of the network. Finally, Scenario $4$ shows a validation loss between the other Scenarios. As we will see later, the network is able to find gender features in the face embeddings, even if the network and the embeddings were not trained for gender recognition. As we can see in Figure~\ref{val_loss}, the validation loss obtained with biased scores and sensitive features (Scenario $2$) is lower than the validation losses obtained for biased scores and blind features (Scenarios $3$ and $4$).

\begin{figure}[t]
\centering
\includegraphics[width=0.95\columnwidth]{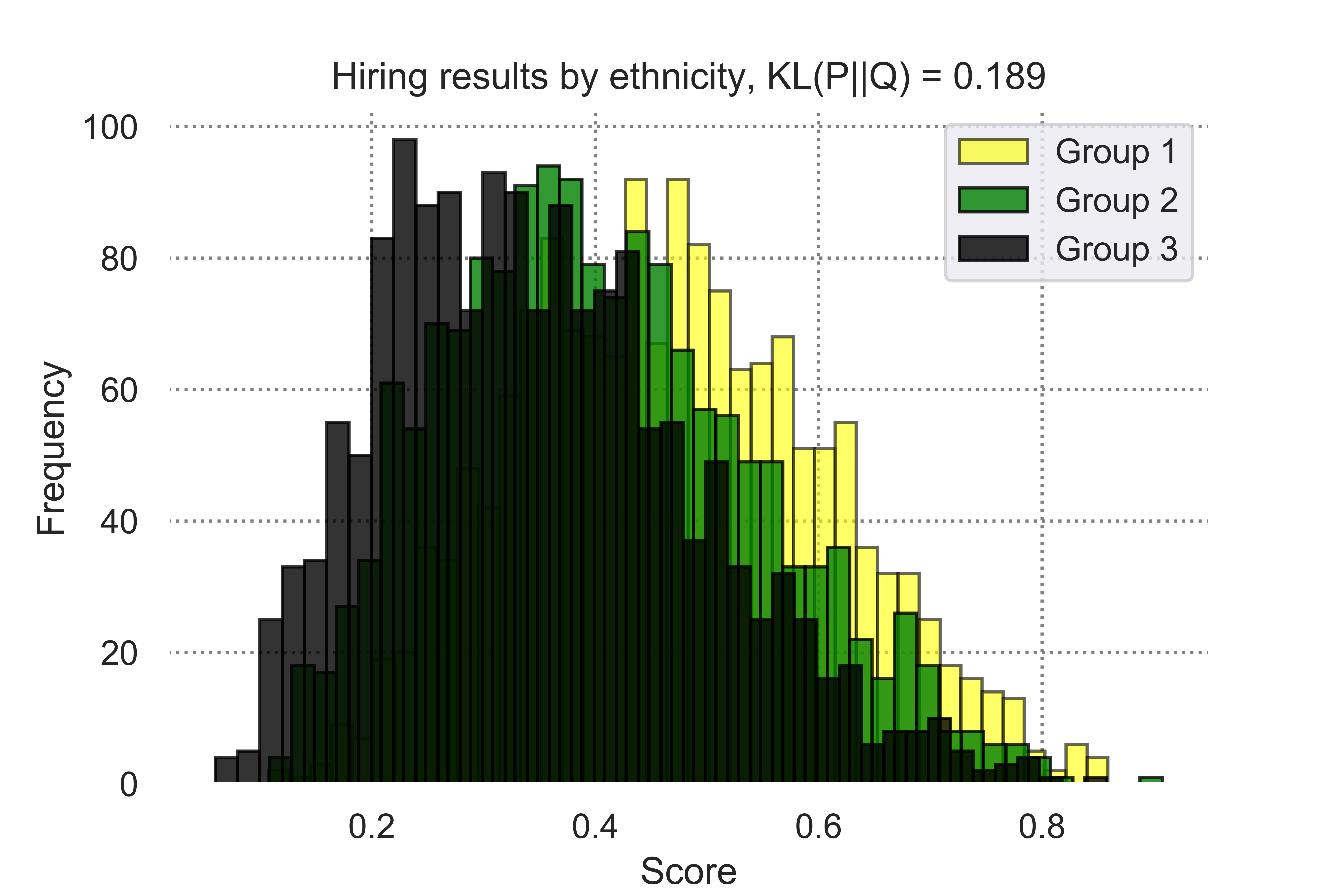} 
\caption{Hiring score distributions by ethnicity group trained according to the setup of the Scenario 4.}
\label{ethnicity_bias}
\end{figure}

In Figure~\ref{gender_distribution} we can see the  distributions of the scores predicted in each scenario by gender, where the presence of the bias is clearly visible in some plots. For each scenario, we compute the Kullback-Leibler divergence $\textrm{KL}(P||Q)$ from the female score distribution $Q$ to the male $P$ as a measure of the bias' impact on the classifier output. In Scenarios $1$ and $3$, Figure \ref{gender_distribution}.a and \ref{gender_distribution}.c respectively, there is no gender difference in the scores, a fact that we can corroborate with the KL divergence tending to zero (see top label in each plot). In the first case (Scenario $1$) we obtain those results because we used the unbiased scores $T^U$ during the training, so that the gender information in the input becomes irrelevant for the model, but in the second one (Scenario $3$) because we made sure that there was no gender information in the training data, and both classes were balanced. Despite using a target function biased, the absence of this information makes the network blind to this bias, paying this effect with a drop of performance with respect to the gender-biased scores $T^G$, but obtaining a fairer model.

The Scenario $2$ (Figure \ref{gender_distribution}.b) leads us to the model with the most notorious difference between male-female classes (note the KL divergence rising to $0.452$), which makes sense because we're explicitly providing it with gender information. In Scenario $4$ the network is able to detect the gender information from the face embeddings, as mentioned before, and find the correlation between them and the bias injected to the target function. Note that these embeddings were generated by a network originally trained to perform face recognition, not gender recognition. Similarly, gender information could be present in the feature embeddings generated by networks oriented to other tasks (e.g. sentiment analysis, action recognition, etc.). Therefore, despite not having explicit access to the gender attribute, the classifier is able to reproduce the gender bias, even though the attribute gender was not explicitly available during the training (i.e. the gender was inferred from the latent features present in the face image). In this case, the KL divergence is around $0.171$, a lower value than the $0.452$ of Scenario $2$, but anyway ten times higher than Unbiased Scenarios.

Moreover, gender information is not the only sensitive information that algorithms like face recognition models can extract from unstructured data. In Figure~\ref{ethnicity_bias} we present the distributions of the scores by ethnicity predicted by a network trained with Ethnicity-biased scores $T^E$ in an analogous way to Scenario $4$ in the gender experiment. The network is also capable to extract the ethnicity information from the same facial feature embeddings, leading to an ethnicity-biased network when trained with skewed data. In this case, we compute the KL divergence by making $1$-to-$1$ combinations (i.e. G$1$ vs G$2$, G$1$ vs G$3$, and G$2$ vs G$3$) and reporting the average of the three divergences.

\subsection{FairCVtest: Training fair models}
\label{removing bias}

As we have seen, using data with  biased labels is not a big concern if we can assure that there's no information correlated with such bias in the algorithm's input, but we can't always assure that. Unstructured data are a rich source of sensitive information for complex deep learning models, which can exploit the correlations in the dataset, and end up generating undesired discrimination.

\begin{figure}[!t]
    \centering
    \begin{subfigure}[b]{\columnwidth}
         \centering
         \includegraphics[width=\columnwidth]{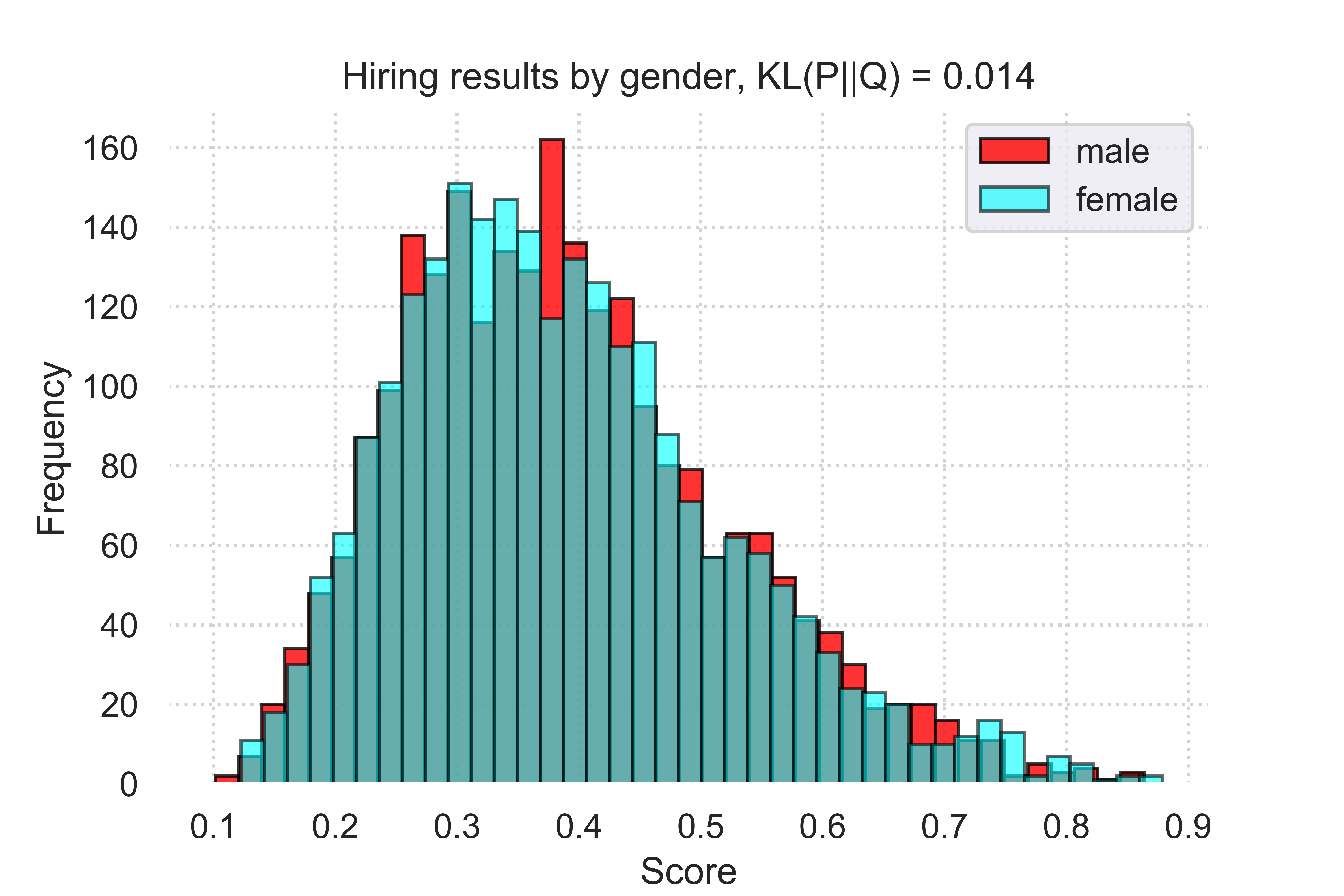}
     \end{subfigure}
    \hspace{-1cm}
    \begin{subfigure}[b]{\columnwidth}
         \centering
         \includegraphics[width=\columnwidth]{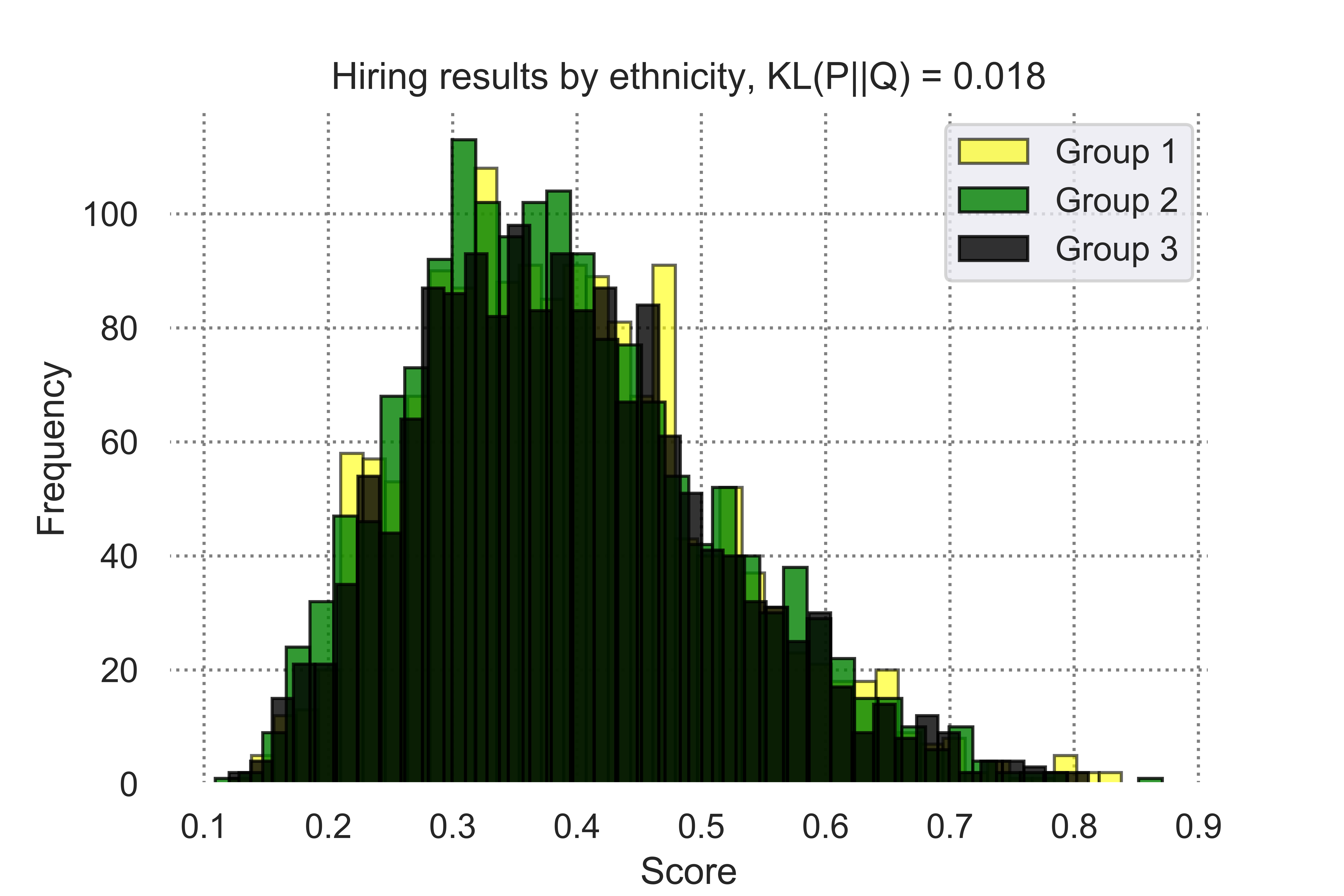}
     \end{subfigure}
    \caption{Hiring score distributions by gender (Up) and ethnicity (Down) after removing sensitive information from the face feature embeddings.}
    \label{bias_rm}
\end{figure}

\setlength{\tabcolsep}{7pt}
\renewcommand{\arraystretch}{1.0}
\begin{table*}[!t]
\begin{center}
\caption{Distribution of the top $100$ candidates for each scenario in FairCVtest, by gender and ethnicity group. $\Delta$ = maximum difference across groups. Dem = Demographic attributes (gender and ethnicity).}
\label{Tab:Best_scores}
\begin{tabular}{|c|c|c|c|c|c|c|c|c|c|c|c|}
\hline
\multirow{2}{*}{\textbf{Scenario}}&\multirow{2}{*}{\textbf{Bias}}&\multicolumn{3}{c|}{\textbf{Input Features}}&\multicolumn{2}{c|}{\textbf{Gender}}&\multirow{2}{*}{\textbf{$\Delta$}}&\multicolumn{3}{c|}{\textbf{Ethnicity}}&\multirow{2}{*}{\textbf{$\Delta$}}\\
\cline{3-7}\cline{9-11}
&&\textbf{Merits}&\textbf{Dem}&\textbf{Face}&\textbf{Male}&\textbf{Female}&&\textbf{Group $1$}&\textbf{Group $2$}&\textbf{Group $3$}&\\
\hline
\hline
$1$ & no & yes & yes & no & $51 \%$&$49 \%$&  $2\%$ &$33 \%$&$34 \%$&$33 \%$&$ 1\%$\\
\hline
$2$ & yes & yes & yes & no & $87 \%$&$13 \%$&$ 74\%$&$90 \%$&$9 \%$&$1 \%$&$ 89\%$\\
\hline
$3$ & yes & yes & no & no &$50 \%$&$50 \%$&$ 0\%$ &$32 \%$&$34 \%$&$34 \%$&$  2\%$\\
\hline
$4$ & yes & yes & no & yes &$77 \%$&$23 \%$&$ 54\%$ &$53 \%$&$31 \%$&$16 \%$&$ 37\%$\\
\hline
Agnostic & yes & yes & no & yes &$50 \%$&$50 \%$&$ 0\%$ &$ 35 \%$&$30 \%$&$35 \%$&$ 5\%$\\
\hline

\end{tabular}
\end{center}
\end{table*}
\setlength{\tabcolsep}{1.4pt}

Removing all sensitive information from the input in a general AI setup is almost infeasible, e.g. \cite{Bias_in_Bios} demonstrates how removing explicit gender indicators from personal biographies is not enough to remove the gender bias from an occupation classifier, as other words may serve as ``proxy''. On the other hand, collecting large datasets that represent broad social diversity in a balanced manner can be extremely costly. Therefore, researchers in AI and machine learning have devised various ways to prevent algorithmic discrimination when working with unbalanced datasets including sensitive data. 
Some works in this line of fair AI propose methods that act on the decision rules (i.e. algorithm's output) to combat discrimination \cite{Exploring_discrimination,Target_gaussian}.In \cite{FairnessGAN} the authors develop a method to generate synthetic datasets that approximate a given original one, but more fair with respect to certain protected attributes. Other works focus on the learning process as the key point to prevent biased models. The authors of \cite{Right_reason} propose an adaptation of DANN \cite{DANN}, originally proposed to perform domain adaptation, to generate agnostic feature representations, unbiased related to some protected concept. In \cite{Bias_Bios_protected} the authors propose a method to mitigate bias in occupation classification without having access to protected attributes, by reducing the correlation between the classifier's output for each individual and the word embeddings of their names. A joint learning and unlearning method is proposed in \cite{Turning_Blind_Eye} to simultaneously learn the main classification task while unlearning biases by applying confusion loss, based on computing the cross entropy between the output of the best bias classifier and an uniform distribution. The authors of \cite{Learning_not_to_Learn} propose a new regularization loss based on mutual information between feature embeddings and bias, training the networks using adversarial \cite{GAN} and gradient reversal \cite{DANN} techniques. Finally, in \cite{SensitiveNets} an extension of triplet loss \cite{TripletLoss} is applied to remove sensitive information in feature embeddings, without losing performance in the main task.

In this work we have used the method proposed in \cite{SensitiveNets} to generate agnostic representations with regard to gender and ethnicity information. This method was proposed to improve privacy in face biometrics by incorporating an adversarial regularizer capable of removing the sensitive information from the learned representations, see \cite{SensitiveNets} for more details. The learning strategy is defined in this case as:

\begin{equation}
\label{eqn:learning_sensitivenets}
     \min_{\textbf{w}}{\sum_{\textbf{x}^j\in \mathcal{S}}(\mathcal{L}(O(\textbf{x}^j|\textbf{w}),T^j)+\Delta^j)} 
\end{equation}
where  $\Delta^j$  is generated with a sensitiveness detector and measures the amount of sensitive information in the learned model represented by $\mathbf{w}$. We have trained the face representation used in the Scenario $4$ according to this method (named as Agnostic scenario in next experiments).

In Figure~\ref{bias_rm} we present the distributions of the hiring scores predicted using the new agnostic embeddings for the face photographs instead of the previous ResNet-$50$ embeddings (Scenario $4$, compare with Figure \ref{gender_distribution}.d). As we can see, after the sensitive information removal the network can't extract gender information from the embeddings. As a result, the two distributions are balanced despite using the gender-biased labels and facial information. In Figure~\ref{bias_rm} we can see the results of the same experiment using the ethnicity-biased labels (compare with Figure \ref{ethnicity_bias}). Just like the gender case, the three distributions are also balanced after removing the sensitive information from the face feature embeddings, obtaining an ethnicity agnostic representation. In both cases the KL divergence shows values similar to those obtained for unbiased Scenarios.

Previous results suggest the potential of sensitive information removal techniques to guarantee fair representations. In order to evaluate further these agnostic representations, we conducted another experiment simulating the outcomes of a recruitment tool. We assume that the final decision in a recruitment process will be managed by humans, and the recruitment tool will be used to realize a first screening among a large list of candidates including the $4$,$800$ resumes used as validation set in our previous experiments. For each scenario, we simulate the candidates screening by choosing the top $100$ scores among them (i.e. scores with highest values). We present the distribution of these selections by gender and ethnicity in Table \ref{Tab:Best_scores}, as well as the maximum difference across groups ($\Delta$). As we can observe, in Scenarios $1$ and $3$, where the classifier shows no demographic bias, we have almost no difference $\Delta$ in the percentage of candidates selected from each demographic group. On the other hand, in Scenarios $2$ and $4$ the impact of the bias is notorious, being larger in the first one with a difference of $74\%$ in the gender case and $89\%$ in the ethnicity case. The results show differences of $54\%$ for the gender attribute in the Scenario $4$, and $37\%$ for the ethnicity attribute. However, when the sensitive features removal technique is applied \cite{SensitiveNets}, the demographic difference drops from $54\%$ to $0\%$ in the gender case, and from $37\%$ to $5\%$ in the ethnicity one, effectively correcting the bias in the dataset. These results demonstrate the potential hazards of these recruitment tools in terms of fairness, and also serve to show possible ways to solve them.

\section{Conclusions}\label{conclusions}

We present FairCVtest, a new experimental framework (publicly available\footnote{\url{https://github.com/BiDAlab/FairCVtest}}) on AI-based automated recruitment to study how multimodal machine learning is affected by biases present in the training data. Using FairCVtest, we have studied the capacity of common deep learning algorithms to expose and exploit sensitive information from commonly used structured and unstructured data.

The contributed experimental framework includes FairCVdb, a large set of $24$,$000$ synthetic profiles with information typically found in job applicants' resumes. These profiles were scored introducing gender and ethnicity biases, which resulted in gender and ethnicity discrimination in the learned models targeted to generate candidate scores for hiring purposes. Discrimination was observed not only when those gender and ethnicity attributes were explicitly given to the system, but also when a face image was given instead. In this scenario, the system was able to expose sensitive information from these images (gender and ethnicity), and model its relation to the biases in the problem at hand. This behavior is not limited to the case studied, where bias lies in the target function. Feature selection or unbalanced data can also become sources of biases. This last case is common when datasets are collected from historical sources that fail to represent the diversity of our society.

Finally, we discussed recent methods to prevent undesired effects of these biases, and then experimented with one of these methods (SensitiveNets) to improve fairness in this AI-based recruitment framework. Instead of removing the sensitive information at the input level, which may not be possible or practical, SensitiveNets removes sensitive information during the learning process. 


The most common approach to analyze algorithmic discrimination is through group-based bias \cite{serna2020formulation}. However, recent works are now starting to investigate biased effects in AI with user-specific methods, e.g. \cite{pentland2020fair,varshney2020AIfairness}. Future work will update FairCVtest with such user-specific biases in addition to the considered group-based bias.

\section{ Acknowledgments}
\label{ack}
This work has been supported by projects BIBECA (RTI$2018$-$101248$-B-I$00$ MINECO/FEDER), TRESPASS-ETN (MSCA-ITN-$2019$-$860813$), PRIMA (MSCA-ITN-$2019$-$860315$); and by Accenture. A. Peña is supported by a research fellowship from Spanish MINECO.

\clearpage
{\small
\bibliographystyle{ieee_fullname}
\bibliography{egbib}
}

\newcommand{\comment}[1]{}

\comment{
\setlength{\tabcolsep}{4pt}
\begin{table*}[!t]
\begin{center}
\caption{.}
\label{Tab:Best_scores}
\begin{tabular}{|c|l|c|c|c|c|c|c|c|}
\hline
\multirow{2}{*}{Scenario}&\multirow{2}{*}{Candidates}&\multicolumn{2}{c|}{Gender}&\multirow{2}{*}{SER}&\multicolumn{3}{c|}{Ethnicity}&\multirow{2}{*}{SER}\\
\cline{3-4}\cline{6-8}
&&Male&Female&&Group 1&Group 2&Group 3&\\
\hline
\hline
\multirow{4}{*}{1}&Best 10&60 \%&40 \%&&40 \%&30 \%&30 \%&\\
\cline{2-9}
&Best 50&48 \%&52 \%&&38 \%&36 \%&36 \%&\\
\cline{2-9}
&Best 100&52 \%&48 \%&&34 \%&33 \%&33 \%&\\
\cline{2-9}
&Best 200&48.5 \%&51.5 \%&&32 \%&34.5 \%&33.5 \%&\\
\hline
\multirow{4}{*}{2}&Best 10&100 \%&0 \%&&90 \%&10 \%&0 \%&\\
\cline{2-9}
&Best 50&88 \%&12 \%&&78 \%&16 \%&6 \%&\\
\cline{2-9}
&Best 100&87 \%&13 \%&&77 \%&19 \%&4 \%&\\
\cline{2-9}
&Best 200&83.5 \%&16.5 \%&&75.5 \%&17.5 \%&7 \%&\\
\hline
\multirow{4}{*}{3}&Best 10&50 \%&50 \%&&30 \%&30 \%&40 \%&\\
\cline{2-9}
&Best 50&44 \%&56 \%&&36 \%&28 \%&36 \%&\\
\cline{2-9}
&Best 100&50 \%&50 \%&&37 \%&31 \%&32 \%&\\
\cline{2-9}
&Best 200&49 \%&51 \%&&36 \%&31.5 \%&32.5 \%&\\
\hline
\multirow{4}{*}{4}&Best 10&90 \%&10 \%&&70 \%&30 \%&0 \%&\\
\cline{2-9}
&Best 50&78 \%&22 \%&&56 \%&16 \%&28 \%&\\
\cline{2-9}
&Best 100&77 \%&23 \%&&50 \%&25 \%&25 \%&\\
\cline{2-9}
&Best 200&70.5 \%&29.5 \%&&49 \%&27 \%&24 \%&\\
\hline
\multirow{4}{*}{Agnostic}&Best 10&40 \%&60 \%&&10 \%&40 \%&50 \%&\\
\cline{2-9}
&Best 50&42 \%&58 \%&&28 \%&40 \%&32 \%&\\
\cline{2-9}
&Best 100&50 \%&50 \%&&35 \%&31 \%&34 \%&\\
\cline{2-9}
&Best 200&51.5 \%&48.5 \%&&31 \%&33.5 \%&35.5 \%&\\
\hline

\end{tabular}
\end{center}
\end{table*}
\setlength{\tabcolsep}{1.4pt}
}

\end{document}